\documentclass[letterpaper, 10 pt, conference]{ieeeconf}  
\usepackage{graphicx}
\usepackage[table]{xcolor}
\usepackage{booktabs} 

\IEEEoverridecommandlockouts                              

\title{\LARGE \bf
Saliency-Guided Domain Adaptation for Left-Hand Driving in Autonomous Steering
}

\author{Zahra Mehraban$^{1}$, Sebastien Glaser$^{1}$, Michael Milford$^{2}$, and Ronald Schroeter$^{1}$ 
\thanks{$^{1}$Centre for Accident Research and Road Safety-Queensland, Queensland University of Technology,
Brisbane 4000, Australia;
        {\tt\small z.mehraban@qut.edu.au, sebastien.glaser@qut.edu.au, r.schroeter@qut.edu.au}}%
\thanks{$^{2}$ QUT Centre for Robotics and School of Electrical Engineering and Computer Science, Queensland University of Technology, Brisbane, QLD 4000, Australia. This work received funding from an ARC Laureate Fellowship FL210100156 to MM
        {\tt\small michael.milford@qut.edu.au}}%
}

\usepackage{subfig}   

\begin{document}

\maketitle
\thispagestyle{empty}
\pagestyle{empty}

\begin{abstract}

Domain adaptation is required for automated driving models to generalize well across diverse road conditions. This paper explores a training method for domain adaptation to adapt PilotNet, an end-to-end deep learning-based model, for left-hand driving conditions using real-world Australian highway data. Four training methods were evaluated: (1) a baseline model trained on U.S. right-hand driving data, (2) a model trained on flipped U.S. data, (3) a model pretrained on U.S. data and then fine-tuned on Australian highways, and (4) a model pretrained on flipped U.S. data and then fine-tuned on Australian highways. This setup examines whether incorporating flipped data enhances the model adaptation by providing an initial left-hand driving alignment. The paper compares model performance regarding steering prediction accuracy and attention, using saliency-based analysis to measure attention shifts across significant road regions. Results show that pretraining on flipped data alone worsens prediction stability due to misaligned feature representations, but significantly improves adaptation when followed by fine-tuning, leading to lower prediction error and stronger focus on left-side cues. To validate this approach across different architectures, the same experiments were done on ResNet, which confirmed similar adaptation trends. These findings emphasize the importance of preprocessing techniques, such as flipped-data pretraining, followed by fine-tuning to improve model adaptation with minimal retraining requirements.
\end{abstract}

\section{INTRODUCTION}

Adapting a machine learning model trained on data from one context to perform well in a different dataset is a challenge in artificial intelligence, especially for expanding technologies such as Automated Vehicles (AVs) around the world \cite{bai2024bridging}. This is important when the training environment is significantly different from the deployment environment, such as changes in traffic rules, road layouts, or weather conditions \cite{zhu2023learning}. For instance, models trained on right-hand driving data (e.g., U.S. roads) may struggle when applied to left-hand driving regions (e.g., Australian roads) due to reversed lane positions, road structures, signage, and driving conventions. Such domain shifts can result in poor performance, requiring effective adaptation strategies that adapt existing models without requiring significant amounts of new data or extensive computing resources \cite{farahani2021brief}.
A compelling example of this broader issue is adapting AVs trained on U.S. data for Australian highways. While training a model from scratch on left-hand driving datasets is feasible, domain adaptation offers significant advantages by building on knowledge from larger right-hand driving datasets, reducing computational costs, and potentially improving generalization by retaining useful features learned from the source domain \cite{wang2018deep}. 
\begin{figure}[!t]
    \centering
    \includegraphics[width=\columnwidth]{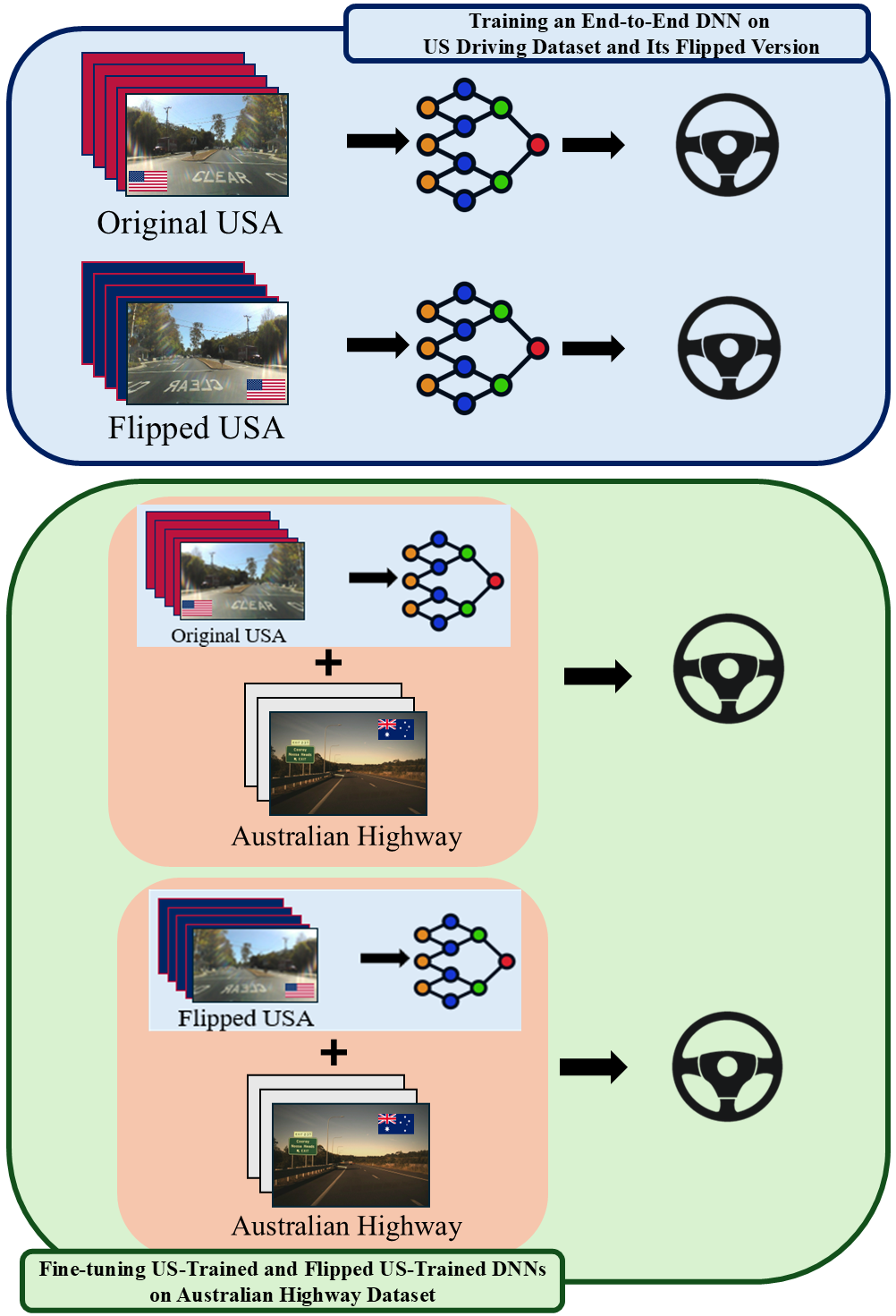}
    \caption{Training Methodology: Pretraining on original/flipped U.S. data (top) and subsequent fine-tuning on Australian highway data (bottom)}
    \label{fig:yourlabel}
\end{figure}
 
A widely used domain adaptation approach in deep learning is fine-tuning \cite{yosinski2014transferable}, where a pretrained model is further trained on target domain data to improve model generalization under various conditions \cite{lester2021power}. However, fine-tuning alone may not be sufficient for successful adaptation when the original training environment is substantially different from the target domain, such as from right-hand to left-hand driving. 

To address this, this paper explores a domain adaptation strategy: pretraining on flipped U.S. data, i.e., flipping images horizontally to mimic left-hand driving conditions, and then fine-tuning on Australian highway data. This method has a preliminary alignment with left-hand driving, which can improve steering prediction accuracy and shift the model's attention toward left-side road features. We analyze whether this flipping-based strategy improves adaptation over direct fine-tuning on end-to-end deep neural networks (DNNs) like PilotNet \cite{bojarski2016end} and ResNet \cite{he2016deep}, which have revolutionized automated driving by enabling vehicles to predict steering angles directly from raw sensor inputs \cite{wadekar2021towards}.

Analyzing the impact of flipping before fine-tuning requires more than conventional metrics such as loss values or steering errors. To address this, attention shifts are represented using saliency maps \cite{gomez2023computing}, which illustrate how models generalize to left-hand driving. The proposed method is evaluated through steering prediction and attention analysis.
The key contributions are:
1) Employing a local dataset collected on Australian highways to assess left-hand driving adaptation.
2) Investigating whether pretraining on flipped U.S. data provides an initial left-hand driving alignment improves fine-tuning efficiency and generalization, by comparing four training strategies (Fig. 1).
3) Using saliency maps to analyze attention shifts across key road regions (left, center, right), and to reveal how flipping before fine-tuning helps models focus on left-side road elements.

The study explores the following research questions:
\begin{itemize}

\item How does fine-tuning on Australian highways improve a right-hand trained automated driving model?
\item Does pretraining on flipped U.S. data improve adaptation compared to direct fine-tuning?
\item How do training strategies influence the attention distribution of the model in saliency maps?
\end{itemize}

\section{RELATED WORKS}

\subsection{Domain Adaptation}

Automated driving models often struggle to generalize across different road environments because of changes in road structures, weather conditions, and traffic rules. This challenge, known as the domain gap, arises when a model trained on one dataset (source domain) performs poorly on another dataset (target domain) due to shifts in distribution \cite{farahani2021brief}. Domain adaptation methods are regularly used to address this problem by transferring knowledge from the source domain to the target domain so that deep learning models can generalize better in unseen environments. Various approaches have been proposed, including feature alignment \cite{sun2016deep}, image style transfer \cite{hoffman2018cycada}, adversarial training \cite{tzeng2017adversarial}, and self-training \cite{liu2021cycle}. 

In the context of AVs, domain adaptation has been used to transfer knowledge between simulated and real-world driving environments \cite{bousmalis2017unsupervised},  between different geographic locations \cite{zhu2023learning}, from daytime to night driving conditions \cite{ye2022unsupervised, wu2021dannet}, and between several other contextual variations \cite{chen2018domain,ciampi2021domain}. However, limited research has addressed domain adaptation for left-hand driving environments, where road structures and driving directions differ significantly. This work contributes to bridging this gap by introducing a simple flipped-data method before fine-tuning as an adaptation approach for left-hand driving conditions.

Fine-tuning, a transfer learning technique, adapts a pretrained model to a new domain by updating its parameters with target data \cite{5288526}. It is particularly useful when the target domain has limited labeled data, as it allows models to retain knowledge from the source domain while gradually adapting to new driving conditions. Prior studies have shown that fine-tuning improves steering prediction accuracy \cite{hong2024deep}, enhances robustness against environmental changes such as lighting variations and road textures \cite{rothmeier2024time}, adapts driving policies across regions \cite{deng2021adapt}, and bridge the gap between simulation and real-world driving \cite{bousmalis2017unsupervised}. 

This paper builds on these findings by exploring a flipped-data method before fine-tuning, which allows the model to pre-align with left-hand driving conditions before exposure to real Australian road data. By evaluating the impact of flipped-data pretraining on model adaptation, this study provides insights into whether simple preprocessing techniques can enhance domain adaptation performance without requiring extensive retraining.

\subsection{Saliency Maps for Explainability in AVs}

Saliency maps visually represent the decision-making process of a neural network by indicating which input features had the most impact \cite{gomez2023computing}. In AVs, they are used to determine whether models pay attention to correct road features such as lane markings, road edges, and vehicles \cite{kim2017interpretable}. Techniques such as CAM \cite{zhou2016learning}, Grad-CAM \cite{selvaraju2017grad}, and Integrated Gradients \cite{sundararajan2017axiomatic} have been widely used to examine model interpretability. Previous work has demonstrated that models trained on specific datasets may exhibit unintended biases, such as focusing on background objects rather than lane boundaries \cite{zhao2024more}. Through saliency-based analysis, this paper explores how flipping data before fine-tuning affects model attention distribution and whether this approach improves alignment with left-hand driving conditions.

Understanding attention shifts is crucial in evaluating how well a model adapts to new environments. Studies have shown that human drivers exhibit distinct attention patterns when switching between different roads \cite{doshi2009roles}. Similarly, AVs trained on one data may struggle to focus on relevant road features in a different driving data. Research by \cite{droste2020unified} highlights the importance of comparing saliency maps before and after adaptation to assess changes in model perception. This research extends this approach by systematically analyzing how model attention shifts after fine-tuning the flipped U.S. model on Australian highway data. The findings provide insights into how simple preprocessing strategies (flipping) combined with fine-tuning can enhance both model accuracy and perceptual adaptation to left-hand driving conditions.

\section{APPROACH}

This paper investigates a domain adaptation strategy to adapt PilotNet to left-hand driving conditions. The primary goal is to determine whether pretraining on flipped right-hand driving data before fine-tuning on local left-hand driving data improves model adaptation. Four training strategies were implemented to evaluate the effectiveness of pretraining on flipped data before fine-tuning: 

\begin{itemize}

\item Training on U.S. Data: PilotNet trained on U.S. data (right-hand driving) and tested directly on Australian highways.
\item Training on Flipped U.S. Data: PilotNet trained on flipped U.S. data (horizontally flipped images) and tested directly on Australian highways, with no fine-tuning on local data. 
\item Fine-tuning the U.S. pretrained: The U.S. pretrained model was fine-tuned using Australian highway data.
\item Fine-tuning the flipped U.S. pretrained: The flipped U.S. pretrained model was fine-tuned using Australian highway data. This strategy aimed to help the model internalize left-hand driving patterns before exposure to real Australian roads.

\end{itemize}
While PilotNet, a relatively simple convolutional neural network, is explored in detail, the same experiments were conducted on ResNet, a deeper and more complex architecture, to validate the findings across different architectures (see Table I for a summary of both PilotNet and ResNet performance). The approach consists of dataset preparation, training strategies, and evaluation through steering performance and saliency analysis.

\section{EXPERIMENTAL SETUP}

In this section, we describe the experimental setup used to evaluate our domain adaptation approach. It includes the information about the local left-hand driving dataset used, training implementation details, and our evaluation metrics with a focus on the steering angle prediction accuracy and attention distribution analysis.

\subsection{Dataset Preparation}

Pretraining models were performed using the NVIDIA dataset, which had already been preprocessed by other researchers and downsampled to 10 FPS \cite{bojarski2016end}. For fine-tuning, our dataset was sampled from a set of large videos collected over a few days at 26 FPS in different road conditions in Queensland, Australia. For a more targeted dataset, five hours and five minutes of Australian highway driving were selected, which excluded the urban and non-highway scenarios. This made the dataset more specific to highway conditions such as straight roads and mild curves. 

Highway conditions were specifically selected for this study as they provide an ideal context for evaluating lane-keeping adaptation between right-hand and left-hand driving systems. The consistent lane structures and road geometries of highways allow for a more controlled analysis of how the model's attention shifts from right to left-side road features, which is the primary focus of this research. 

To avoid concentrating on long sequences of straight roads, final frames were chosen based on the steer variation and not on the frame sequence. This minimized redundancy and resulted in 30,000 diverse frames that are sufficient for a reasonable size of the dataset, but not a vast amount of data for the purpose of training. 

Several preprocessing steps were taken to improve the relevance and model compatibility. First, 400 pixels were cropped from the top of each 1920 × 1200 frame to remove irrelevant regions, such as the sky. The frames were then resized to 200 × 66 to maintain compatibility with the NVIDIA dataset and to reduce computational requirements. They were then normalized to the YUV color space to enable the model to focus on the areas of the road that are most crucial. 

Overall, these procedures, along with the frame selection strategy, enhanced data quality and reduced computational cost, making the dataset more suitable for fine-tuning models for left-hand driving conditions (Fig. 2).

\begin{figure}[!b]
    \centering
    \includegraphics[width=\columnwidth]{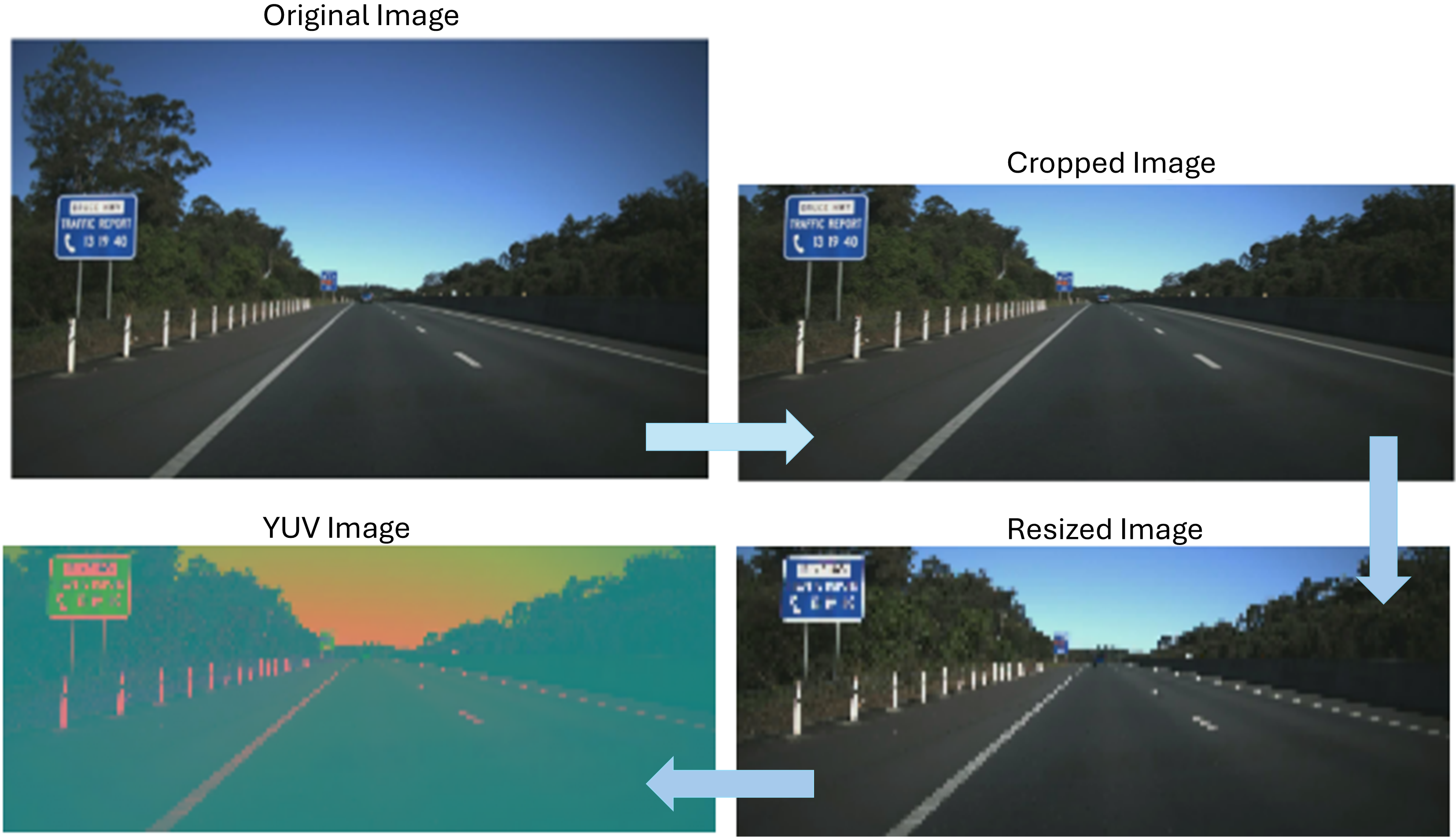}
    \caption{Image preprocessing steps for a random image, including cropping the top 400 pixels to remove irrelevant areas, resizing to 200 × 66 for consistency with the NVIDIA dataset, and converting to YUV color space to improve road feature extraction. }
    \label{fig:yourlabel}
\end{figure}

\subsection{Training Strategies Implementation}

All four models were trained using the same set of hyperparameters to ensure a fair comparison. The pretraining on the U.S. dataset (both original and flipped) used a learning rate of $10^{-4}$, L2 regularization of $10^{-3}$, and batch size of 128 for 50 epochs. For fine-tuning on Australian data, the learning rate was also set to $10^{-4}$ and L2 regularization was reduced to $10^{-4}$. Training was conducted for 15 epochs.
The models were evaluated based on loss values, steering prediction accuracy, and saliency maps. These assessments provided insights into how well the models adapted to Australian highway conditions and how their focus shifted under different training strategies. 

While traditional performance metrics, such as loss values and steering prediction accuracy, indicate overall adaptation, saliency analysis is used to assess how pretraining influences attention distribution across road regions.
The goal is to improve driving model performance for left-hand driving conditions through strategic pretraining on flipped data. This approach leverages the observation that models pretrained on flipped images tend to develop stronger attention patterns toward the left side of the road, which is critical for left-hand driving.

To get a more accurate approximation of attention distribution across models, the mean saliency map was derived from a random subset of test images using Integrated Gradient \cite{sundararajan2017axiomatic}. The approach ensured that the observed attention patterns were not biased by individual images but reflected general trends across different adaptation strategies. To systematically analyze how the models allocated attention, the images were divided into three regions: Left (from left image border to left lane line), Center (between lines), and Right (from right lane line to right image border) (Fig. 3), and saliency spread ratios were computed in these regions.

To ensure that the attention analysis focused on relevant road features, a Region of Interest (ROI) was applied, excluding irrelevant areas such as the sky, trees, and surrounding objects. While all four models had some form of attention in background regions, applying the ROI allowed for a more precise interpretation of attention shifts that directly related to road features. To achieve accurate segmentation of lane boundaries within our selected subset of test images, Canny edge detection \cite{canny1986computational} was applied dynamically for each image in the subset, ensuring that region division was based on actual lane markings rather than manually defined boundaries.  The process provided a structured and unbiased method for evaluating attention shifts induced by different training strategies. The ROI used for saliency analysis and edge-detected images after applying ROI are illustrated in Fig. 4.

\begin{figure}[!b]
    \centering
    \includegraphics[width=\columnwidth]{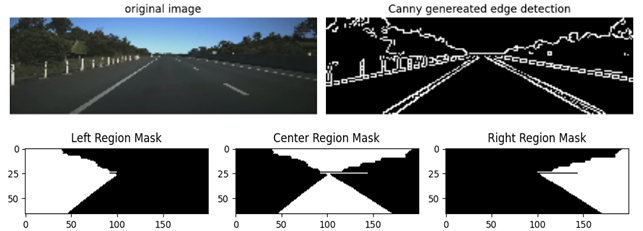}
    \caption{Region masks (Left, Center, Right) generated based on lane boundary detection for a random image to analyze saliency spread across different areas of the road without applying ROI}
    \label{fig:yourlabel}
\end{figure}

\begin{figure}[!t]
    \centering
    \subfloat{\includegraphics[width=\columnwidth]{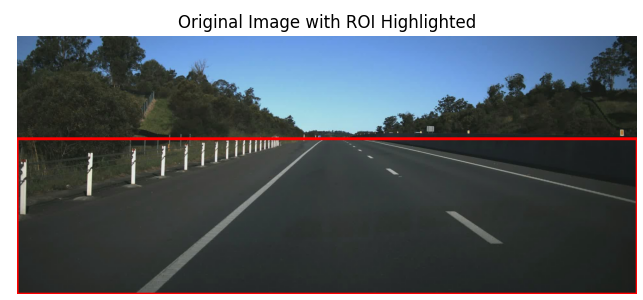}}\\
    \subfloat{\includegraphics[width=\columnwidth]{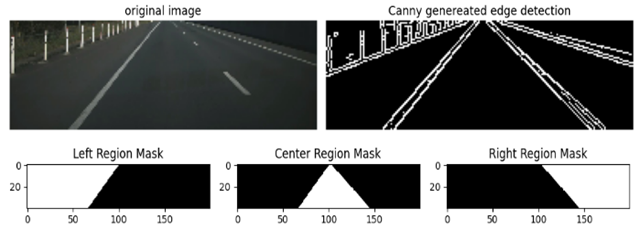}}
    \caption{A random image with ROI highlighted in red and corresponding segmented road regions(Left, Center, Right) after ROI application}
    \label{fig:yourlabel}
\end{figure}

\section{RESULTS AND COMPARISON}

\subsection {Loss Evaluation and Prediction Accuracy}

All four training methods were compared using Mean Squared Error (MSE) to determine their effectiveness in reducing steering prediction error. From Fig. 5, the model pretrained on U.S. data had high MSE, indicating poor generalization performance under Australian road conditions. However, pretraining only on flipped U.S. data resulted in the highest MSE, even worse than the original pretrained model. This poor performance is due to semantic inconsistencies introduced by flipping. For instance, road signs, text-based markers, and other asymmetric features (such as vehicle position relative to lanes) are reflected, potentially confusing the model during inference. Since the model was not trained on correctly oriented left-hand driving data, it acquired incorrect feature associations, and the predictions were random.

Fine-tuning on Australian roads reduced the error significantly, verifying the strength of domain adaptation. However, the best performance came when pretraining on flipped U.S. data was combined with fine-tuning on Australian data. This is because even though pretraining on flipped data added errors, it also provided early exposure to left-hand lane structures, and fine-tuning could effectively correct semantic misinterpretations. Consequently, the flipped-pretrained and fine-tuned model achieved the lowest MSE, providing smoother and more stable steering predictions.

A detailed analysis of predicted outputs, shown in Fig. 6, illustrates the superiority of fine-tuned models over pretrained-only models on original or flipped data. Time-series predictions in Fig. 6 clearly show both fine-tuned models are significantly closer to ground truth steering angles, with deviations of approximately ±5°, compared to the huge oscillations of pretrained-only models. The model trained on the original data had huge deviations of -50° to +40° and could not generalize to left-hand driving conditions, while the model trained on flipped data performed even worse with high instability with steering predictions fluctuating between -60° to +80° and many sudden reversals due to incorrect interpretations of mirrored road signs and reversed lane structures. 

Between the two fine-tuned models, the model pretrained on flipped data showed superior performance. This is particularly evident in the zoomed-in regions, which indicate representative driving segments where this model more accurately tracks the ground truth steering path.

The quantitative superiority of this model is confirmed by the scatter plots in Fig. 6. In these plots, the slope of the linear equation between predicted and actual steering angles indicates how closely the predictions match actual steering needs - a slope closer to 1 means the model predictions more closely match the required steering angles. The linear fit for the fine-tuned flipped data model between predicted and ground truth steering angles has the highest positive slope (0.4399) and the lowest bias (0.0808), indicating the highest correlation with ground truth and lowest systematic errors. Other models, on the other hand, showed lower or even negative correlations, with slopes ranging from -0.47 to 0.22, showing their poorer ability to make stable and accurate steering predictions. 

These results clearly suggest that pretraining on flipped data establishes a good initial left-hand driving direction, and subsequent fine-tuning effectively corrected semantic misinterpretation, leading to improved domain adaptation. 

\begin{figure}[!t]
    \centering
    \includegraphics[width=\columnwidth]{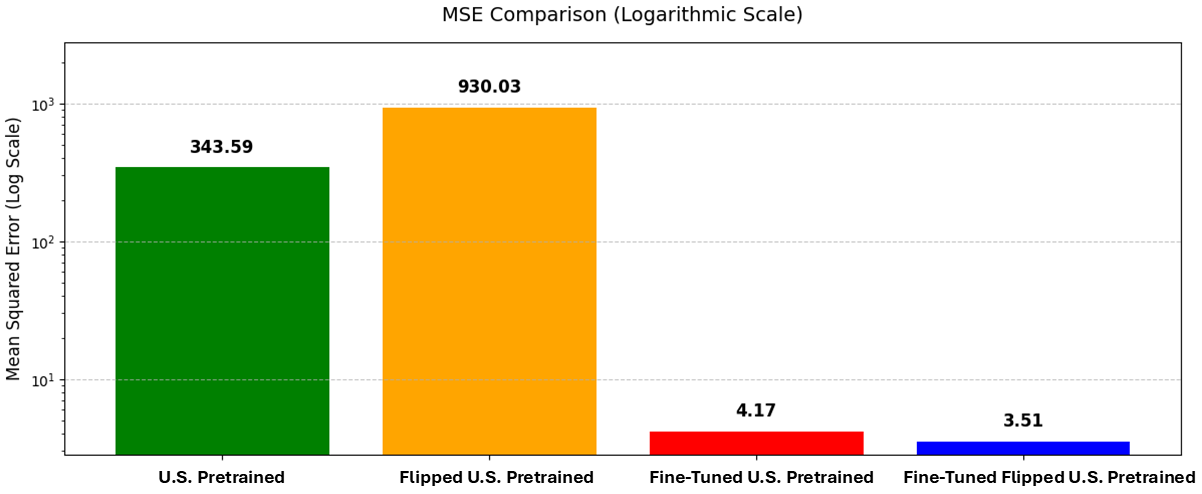}
    \caption{MSE Comparison Across Four Training Strategies - The highest MSE (930.03) from trained on flipped U.S. data versus the lowest MSE (3.51) achieved by combining flipped data pretraining with fine-tuning demonstrates that successful domain adaptation requires both appropriate pretraining and fine-tuning for left-hand driving conditions.}
    \label{fig:yourlabel}
\end{figure}

\begin{figure}[!t]
    \centering
    \subfloat{\includegraphics[width=\columnwidth]{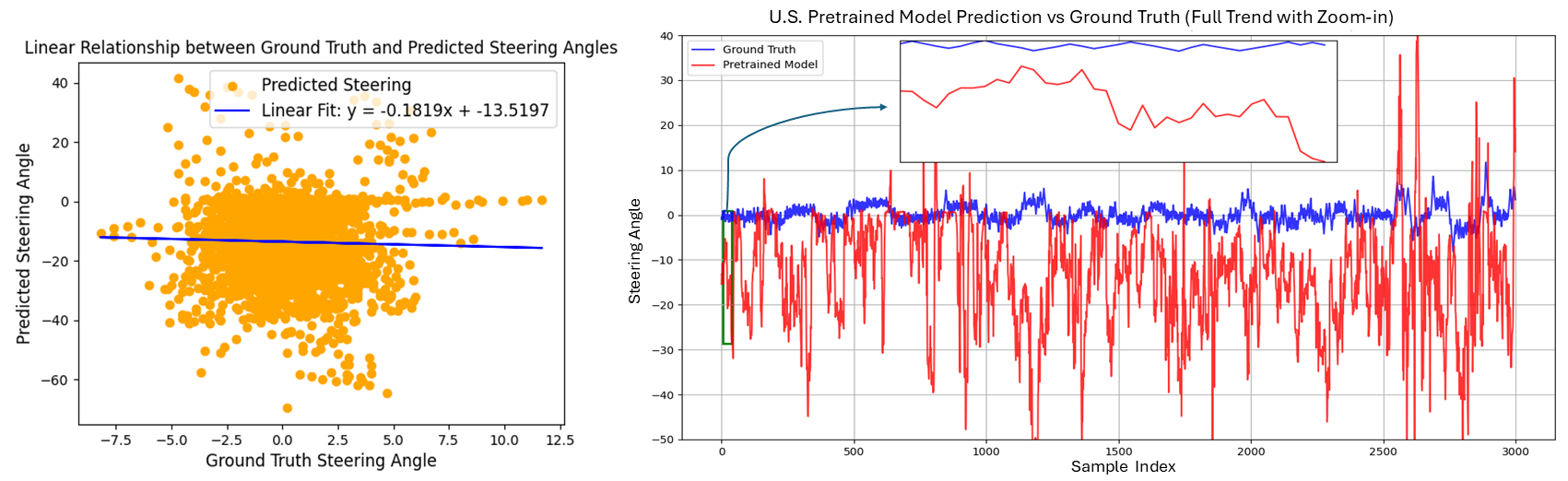}}\\
    \subfloat{\includegraphics[width=\columnwidth]{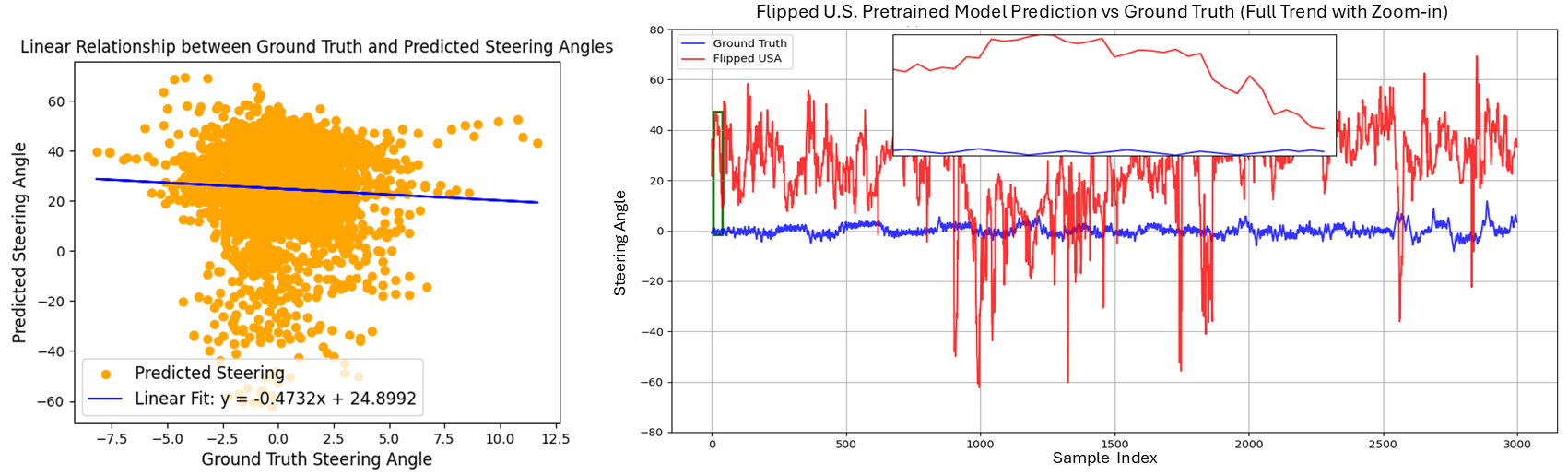}}\\
    \subfloat{\includegraphics[width=\columnwidth]{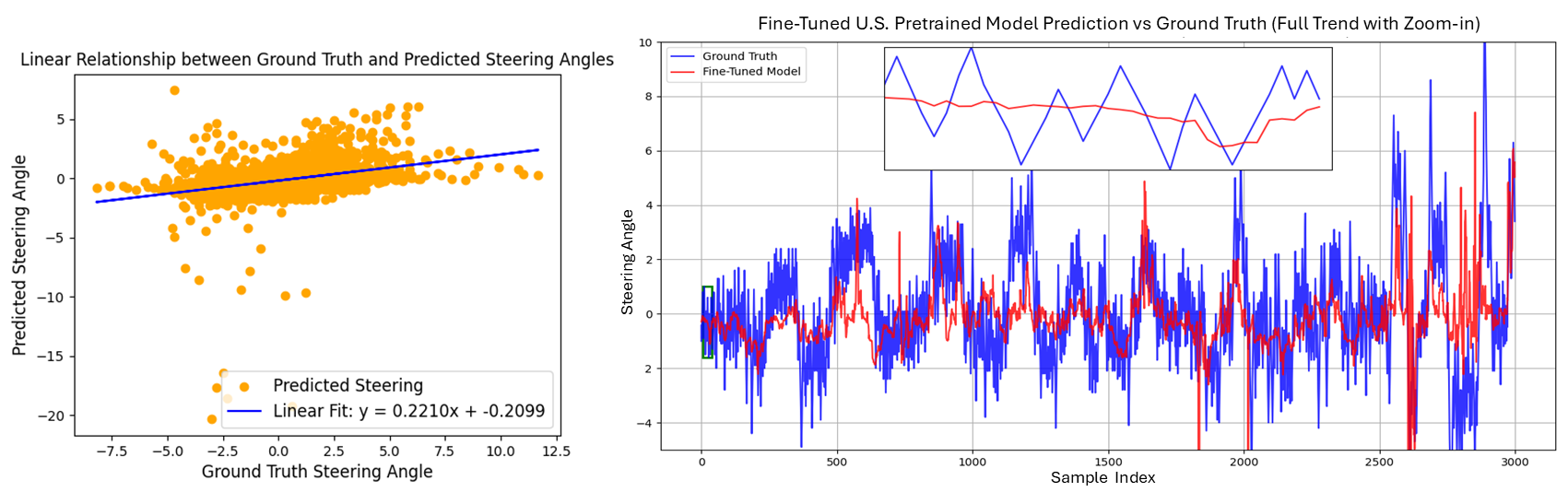}}\\
    \subfloat{\includegraphics[width=\columnwidth]{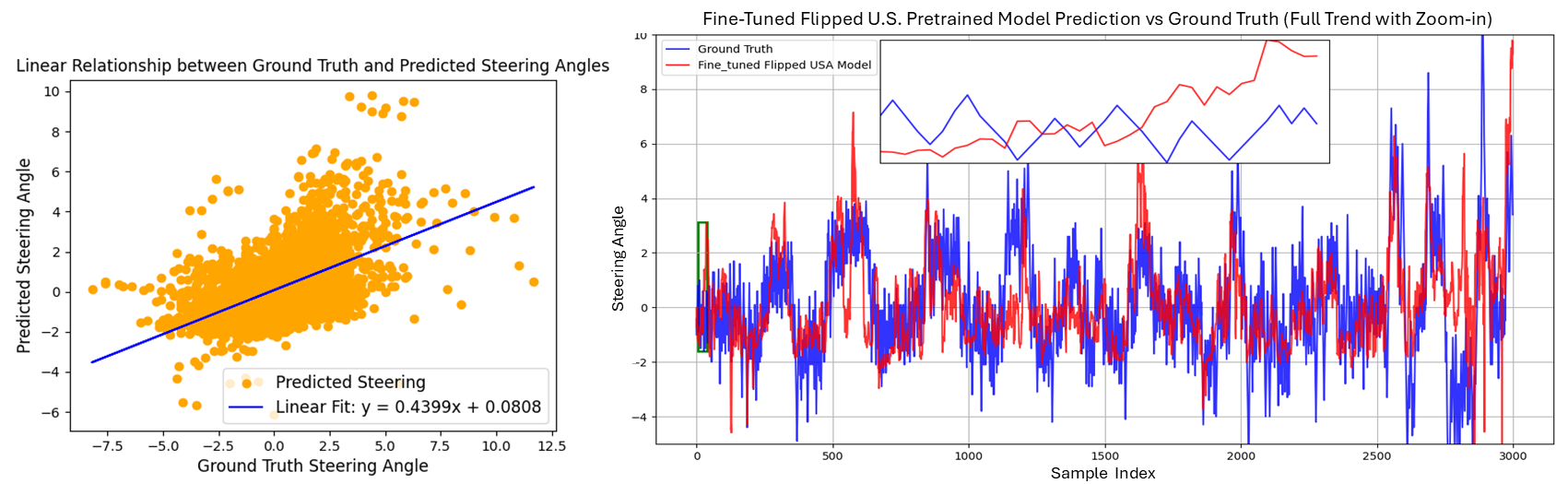}}\\
    \caption{Scatter Plot and Time-Series Comparison of Steering Prediction Performance Across Training Strategies – From top to bottom: U.S. trained model (slope = -0.18), Flipped U.S. trained model (slope= -0.47), fine-tuned U.S. pretrained model (slope = 0.22), and fine-tuned flipped U.S. pretrained model (slope = 0.44) shows progressive improvement in prediction stability and accuracy through scatter plots (left) and time-series tracking (right)}
    \label{fig:yourlabel}
\end{figure}

\subsection {Saliency Map Analysis and Attention Shift}
Saliency maps were analyzed to evaluate how different training strategies influenced the model’s attention to road-relevant features. Instead of relying on pre-defined static areas, a random subset of test images was selected, and for each image, three dynamic regions —left, center, and right —were defined using Canny edge detection to align with actual lane boundaries. The saliency spread (attention distribution) was then computed for each region per image, and the average saliency spread across selected test images was calculated to ensure a more reliable comparison. These average saliency maps are illustrated in Fig. 7. 

As shown in Table I, the U.S. trained model exhibited a strong right-hand bias with $32.5\%$ of its attention on the right, $42.2\%$ in the center, and only $25.3\%$ on the left. This indicates that a model trained from right-hand driving data tends to look at road features relevant to that environment and is not suitable for left-hand driving without adaptation. The fine-tuned model with Australian data proved successful in shifting attention to the left, with $41.16\%$ of attention on the left, showing that fine-tuning alone is effective but not optimal. The flipped U.S. trained model exhibited the most extreme leftward shift, with $74.97\%$ of attention on the left location, $21\%$ in the center, and only $4.42\%$ on the right. While this might seem to be a strong left-hand driving adaptation, this model performed poorly in steering accuracy. The over-left bias can suggest that the model overcompensated on the flipped images without learning proper lane balance, possibly attending to the incorrect features such as flipped signage or distorted road features.	
In contrast, the fine-tuned flipped-pretrained model achieved the most effective adaptation, with $53.35\%$ of attention toward the left, $27\%$ in the center, and $19.2\%$ toward the right.
\begin{figure}[!b]
    \centering
    \includegraphics[width=\columnwidth]{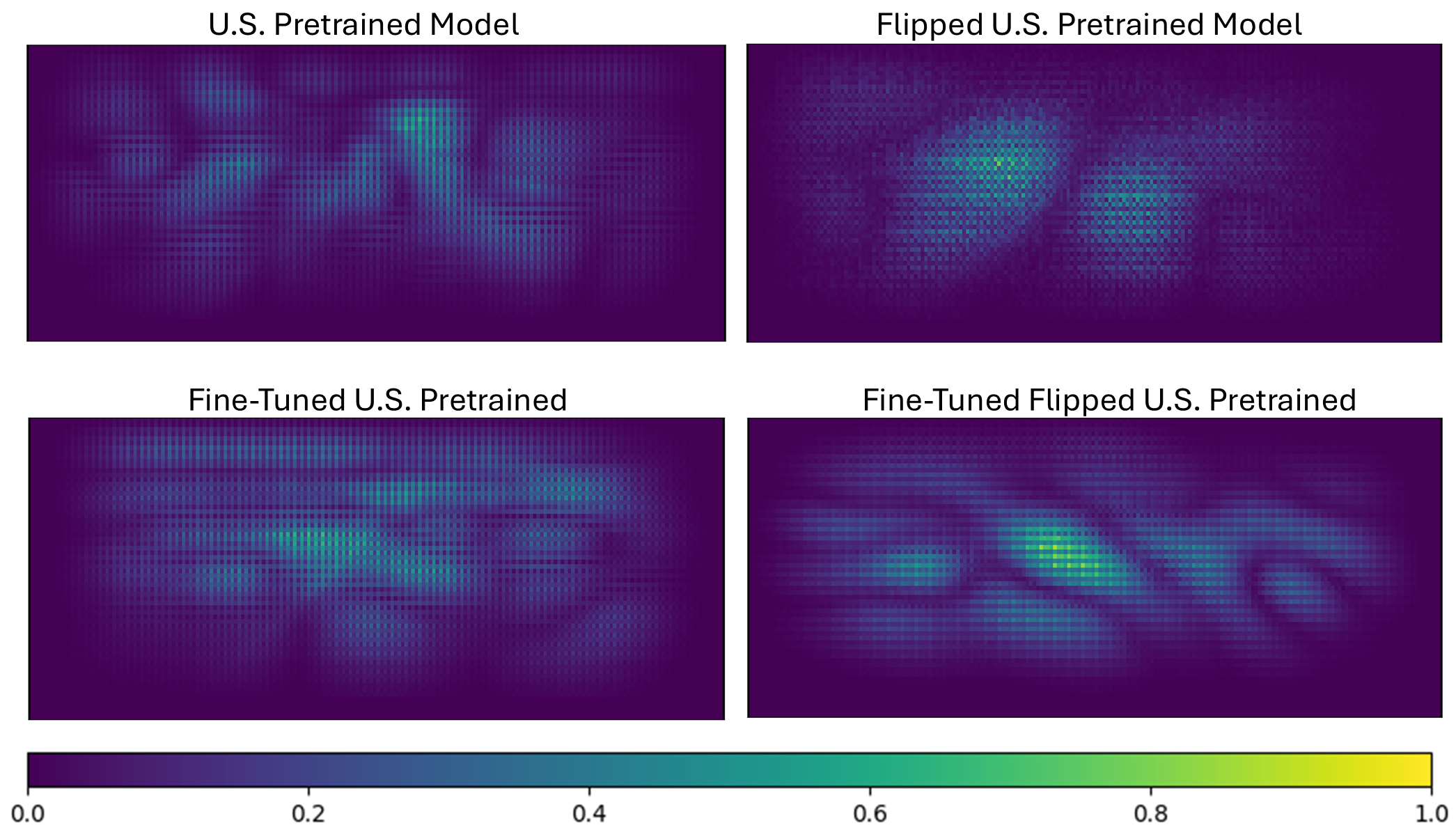}
    \caption{Comparison of Average saliency maps across different training strategies using Integrated Gradient (IG). The intensity represents the attention distribution, highlighting regions of focus for each model. Notably, the Fine-Tuned Flipped U.S. shows a stronger leftward attention shift compared to the others, revealing adaptation to left-hand driving conventions. This shift may influence the model's lane following and maneuvering decision.}
    \label{fig:yourlabel}
\end{figure}

\begin{table*}[t]
    \centering
    \caption{Performance Metric and Saliency Distribution Analysis of Training Strategies for both PilotNet and ResNet}
    \label{tab:performance_comparison}
    \renewcommand{\arraystretch}{0.9} 
    \setlength{\tabcolsep}{6pt} 
    \small 

    \resizebox{\textwidth}{!}{ 
    \begin{tabular*}{\textwidth}{l l c c @{\hspace{1cm}} r r r}
        \toprule[1.3pt]
        \textbf{Model} & \textbf{Training Strategy} & \textbf{Correlation} & \textbf{MSE} & \multicolumn{3}{c}{\textbf{Saliency Distribution (\%)}} \\
        \cmidrule(lr){5-7}
        & & & & \textbf{Left} & \textbf{Center} & \textbf{Right} \\
        \midrule[1.2pt]
        \multicolumn{7}{l}{\textbf{PilotNet}} \\
        \midrule
        \textit{Baseline Models} & & & & & & \\
        & Pretrained on U.S. & -0.1819 & 343.59 & 25.3 & 42.2 & 32.5 \\
        & Pretrained on Flipped U.S. & -0.4732 & 930.03 & 74.97 & 21 & 4.42 \\
        \midrule
        \textit{Fine-Tuning Variations} & & & & & & \\
        & Fine-Tuned (Original) & 0.2210 & 4.17 & 41.16 & 48.46 & 10.3 \\
        & \textbf{Fine-Tuned (Flipped) (Proposed)} & \textbf{0.4399} & \textbf{3.51} & \textbf{53.35} & \textbf{27} & \textbf{19.2} \\
        \midrule[1.3pt]
        \multicolumn{7}{l}{\textbf{ResNet}} \\
        \midrule
        \textit{Baseline Models} & & & & & & \\
        & Pretrained on U.S. & 0.1234 & 359.57 & 16.14 & 32.08 & 51.77 \\
        & Pretrained on Flipped U.S. & -0.5468 & 824.27 & 64.92 & 15.9 & 19.17 \\
        \midrule
        \textit{Fine-Tuning Variations} & & & & & & \\
        & Fine-Tuned (Original) & 0.5229 & 4.18 & 43.95 & 27.1 & 28.93 \\
        & \textbf{Fine-Tuned (Flipped) (Proposed)} & \textbf{0.5538} & \textbf{3.88} & \textbf{52.19} & \textbf{27.94} & \textbf{19.85} \\
        
        \bottomrule[1.3pt]
    \end{tabular*}
    } 
\end{table*}

While still prioritizing left-hand driving features, it maintained some focus on the center and right, ensuring better generalization. Maintaining some level of attention to the center and right might contribute to better lane-following and situational awareness in AVs, and an excess of leftward attention (such as in the flipped-only model) could lead the model to not be able to follow lanes and detect vehicles on adjacent lanes.

\begin{figure}[!b]
    \centering
    \subfloat{\includegraphics[width=\columnwidth]{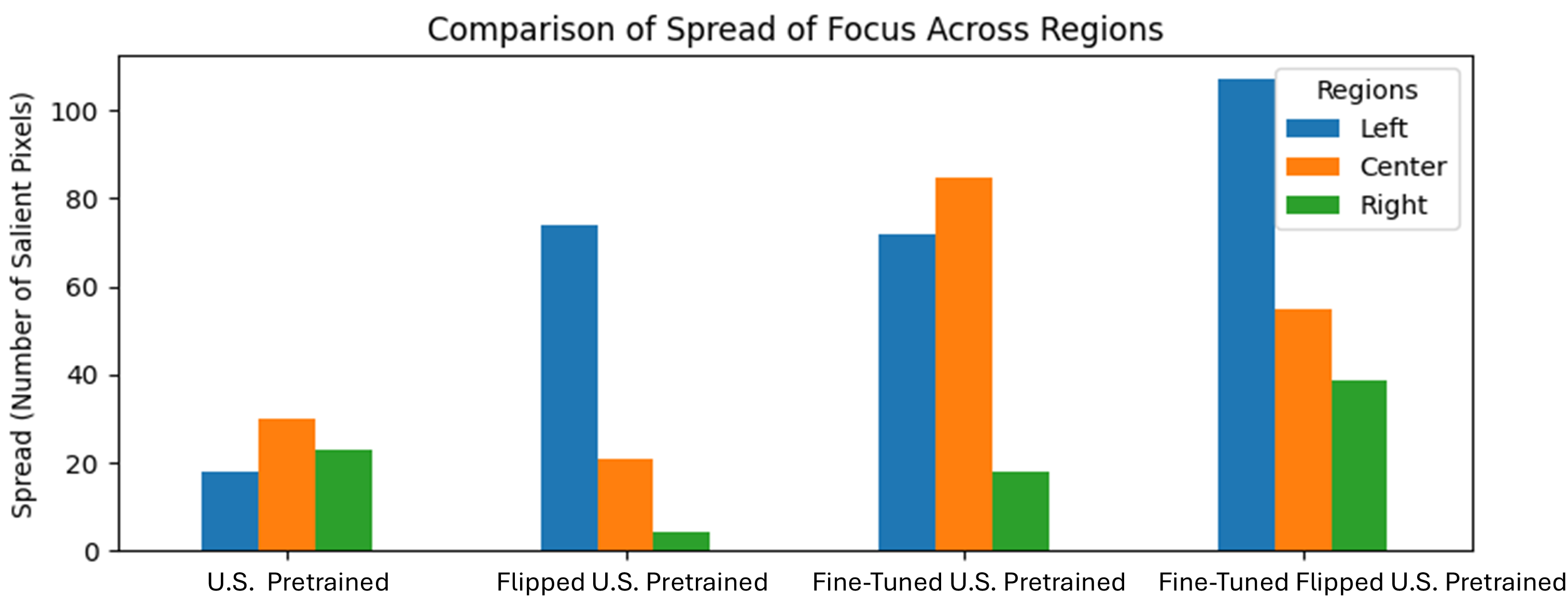}}\\
    \subfloat{\includegraphics[width=\columnwidth]{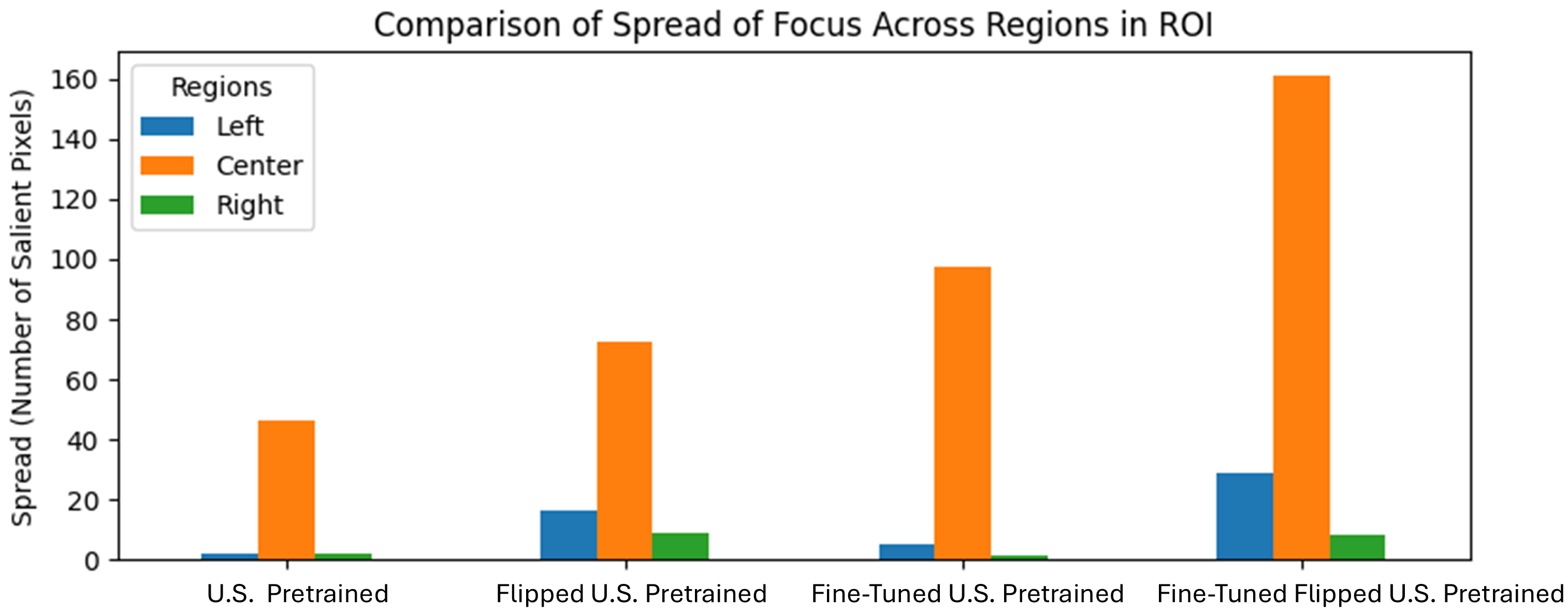}}
    \caption{Saliency distribution comparison across regions before (top) and after ROI application (bottom), demonstrating maintained left-side attention in fine-tuned flipped model even after ROI filtering, indicating genuine road-relevant focus}
    \label{fig:yourlabel}
\end{figure}

To provide a comprehensive description of attention allocation, Fig. 8 illustrates the saliency spread across different road regions under two conditions: (a) without ROI, when all the image regions are considered, and (b) with ROI, when road-related regions are only considered, as shown in Fig.4. The comparison ensures that when we state the model shifts its attention toward the left, this shift is not only occurring in the whole image but also in the critical driving regions. The ROI-based analysis refines the evaluation by isolating the road area, confirming that the leftward shift in attention is not an artifact of background elements but a meaningful adaptation to left-hand driving. This verifies that the fine-tuned flipped model puts the right focus on left-side road details while also finding a proper balance with center-lane attention for safe and stable driving behavior.

To verify whether our domain adaptation method generalizes across architectures, we repeated the experiments with ResNet. This ensures that the observed improvements are applicable to various deep learning architectures rather than being unique to a single model. Table I presents the results, illustrating that ResNet is also exhibiting the same trend with fine-tuning flipped data yielding the
best in MSE reduction, ground truth correlation, and attention to left-hand driving features.
This confirms the robustness of the adaptation method presented with different network architectures.

\section{DISCUSSION AND CONCLUSION}
This paper provides a domain adaptation method for PilotNet using real-world driving data to improve left-hand driving performance. Four training strategies were tested, showing that training on flipped U.S. data alone led to misinterpretations, while combining it with fine-tuning significantly improved adaptation. To validate generalizability, the same experiments were conducted on ResNet, confirming the same trend. Pretraining on flipped data before fine-tuning reduced the MSE by $15.83\%$ in PilotNet and $7.18\%$ in ResNet compared to fine-tuning the original U.S. pretrained model. Furthermore, attention to left-hand road features increased from $41.16\%$ to $53.35\%$ in PilotNet and $43.95\%$ to $52.19\%$ in ResNet. These results demonstrate that pretraining of flipped data effectively increases the adaptability using real-world data, and reduces domain shift with minimal additional data.

Future work could extend this domain adaptation method for a broader range of driving conditions, such as urban environments, which would increase challenges such as complex intersections, different signage, and diverse traffic interactions. Testing across a wider range of road types would further validate the method’s generalizability for automated driving. However, adapting the model for such variations may require more robust pretraining techniques or additional fine-tuning strategies.

\addtolength{\textheight}{-12cm}   




\section*{ACKNOWLEDGMENT}

This research was undertaken under the QUT Industry Chair In Empathic Machines, in collaboration with and co-funded by industry partner Seeing Machines Ltd. Through participation in the National Industry PhD Program, it has also been co-funded by the Department of Education. This work also received funding from an ARC Laureate Fellowship FL210100156 to MM. The authors acknowledge continued support from the Queensland University of Technology (QUT) through the Centre for Accident Research and Road Safety-Queensland (CARRS-Q) and the QUT Centre for Robotics for providing research infrastructure and data collection support.


\bibliographystyle{IEEEtran}
\bibliography{references}

\begin{thebibliography}{10}
\providecommand{\url}[1]{#1}
\csname url@samestyle\endcsname
\providecommand{\newblock}{\relax}
\providecommand{\bibinfo}[2]{#2}
\providecommand{\BIBentrySTDinterwordspacing}{\spaceskip=0pt\relax}
\providecommand{\BIBentryALTinterwordstretchfactor}{4}
\providecommand{\BIBentryALTinterwordspacing}{\spaceskip=\fontdimen2\font plus
\BIBentryALTinterwordstretchfactor\fontdimen3\font minus \fontdimen4\font\relax}
\providecommand{\BIBforeignlanguage}[2]{{%
\expandafter\ifx\csname l@#1\endcsname\relax
\typeout{** WARNING: IEEEtran.bst: No hyphenation pattern has been}%
\typeout{** loaded for the language `#1'. Using the pattern for}%
\typeout{** the default language instead.}%
\else
\language=\csname l@#1\endcsname
\fi
#2}}
\providecommand{\BIBdecl}{\relax}
\BIBdecl

\bibitem{bai2024bridging}
X.~Bai, Y.~Luo, L.~Jiang, A.~Gupta, P.~Kaveti, H.~Singh, and S.~Ostadabbas, ``Bridging the domain gap between synthetic and real-world data for autonomous driving,'' \emph{Journal on Autonomous Transportation Systems}, vol.~1, no.~2, pp. 1--15, 2024.

\bibitem{zhu2023learning}
R.~Zhu, P.~Huang, E.~Ohn-Bar, and V.~Saligrama, ``Learning to drive anywhere,'' \emph{arXiv preprint arXiv:2309.12295}, 2023.

\bibitem{farahani2021brief}
A.~Farahani, S.~Voghoei, K.~Rasheed, and H.~R. Arabnia, ``A brief review of domain adaptation,'' \emph{Advances in data science and information engineering: proceedings from ICDATA 2020 and IKE 2020}, pp. 877--894, 2021.

\bibitem{wang2018deep}
M.~Wang and W.~Deng, ``Deep visual domain adaptation: A survey,'' \emph{Neurocomputing}, vol. 312, pp. 135--153, 2018.

\bibitem{yosinski2014transferable}
J.~Yosinski, J.~Clune, Y.~Bengio, and H.~Lipson, ``How transferable are features in deep neural networks?'' \emph{Advances in neural information processing systems}, vol.~27, 2014.

\bibitem{lester2021power}
B.~Lester, R.~Al-Rfou, and N.~Constant, ``The power of scale for parameter-efficient prompt tuning,'' \emph{arXiv preprint arXiv:2104.08691}, 2021.

\bibitem{bojarski2016end}
M.~Bojarski, D.~Del~Testa, D.~Dworakowski, B.~Firner, B.~Flepp, P.~Goyal, L.~D. Jackel, M.~Monfort, U.~Muller, J.~Zhang \emph{et~al.}, ``End to end learning for self-driving cars,'' \emph{arXiv preprint arXiv:1604.07316}, 2016.

\bibitem{he2016deep}
K.~He, X.~Zhang, S.~Ren, and J.~Sun, ``Deep residual learning for image recognition,'' in \emph{Proceedings of the IEEE conference on computer vision and pattern recognition}, 2016, pp. 770--778.

\bibitem{wadekar2021towards}
S.~N. Wadekar, B.~J. Schwartz, S.~S. Kannan, M.~Mar, R.~K. Manna, V.~Chellapandi, D.~J. Gonzalez, and A.~E. Gamal, ``Towards end-to-end deep learning for autonomous racing: On data collection and a unified architecture for steering and throttle prediction,'' \emph{arXiv preprint arXiv:2105.01799}, 2021.

\bibitem{gomez2023computing}
T.~Gomez and H.~Mouch{\`e}re, ``Computing and evaluating saliency maps for image classification: a tutorial,'' \emph{Journal of Electronic Imaging}, vol.~32, no.~2, pp. 020\,801--020\,801, 2023.

\bibitem{sun2016deep}
B.~Sun and K.~Saenko, ``Deep coral: Correlation alignment for deep domain adaptation,'' in \emph{Computer vision--ECCV 2016 workshops: Amsterdam, the Netherlands, October 8-10 and 15-16, 2016, proceedings, part III 14}.\hskip 1em plus 0.5em minus 0.4em\relax Springer, 2016, pp. 443--450.

\bibitem{hoffman2018cycada}
J.~Hoffman, E.~Tzeng, T.~Park, J.-Y. Zhu, P.~Isola, K.~Saenko, A.~Efros, and T.~Darrell, ``Cycada: Cycle-consistent adversarial domain adaptation,'' in \emph{International conference on machine learning}.\hskip 1em plus 0.5em minus 0.4em\relax Pmlr, 2018, pp. 1989--1998.

\bibitem{tzeng2017adversarial}
E.~Tzeng, J.~Hoffman, K.~Saenko, and T.~Darrell, ``Adversarial discriminative domain adaptation,'' in \emph{Proceedings of the IEEE conference on computer vision and pattern recognition}, 2017, pp. 7167--7176.

\bibitem{liu2021cycle}
H.~Liu, J.~Wang, and M.~Long, ``Cycle self-training for domain adaptation,'' \emph{Advances in Neural Information Processing Systems}, vol.~34, pp. 22\,968--22\,981, 2021.

\bibitem{bousmalis2017unsupervised}
K.~Bousmalis, N.~Silberman, D.~Dohan, D.~Erhan, and D.~Krishnan, ``Unsupervised pixel-level domain adaptation with generative adversarial networks,'' in \emph{Proceedings of the IEEE conference on computer vision and pattern recognition}, 2017, pp. 3722--3731.

\bibitem{ye2022unsupervised}
J.~Ye, C.~Fu, G.~Zheng, D.~P. Paudel, and G.~Chen, ``Unsupervised domain adaptation for nighttime aerial tracking,'' in \emph{Proceedings of the IEEE/CVF conference on computer vision and pattern recognition}, 2022, pp. 8896--8905.

\bibitem{wu2021dannet}
X.~Wu, Z.~Wu, H.~Guo, L.~Ju, and S.~Wang, ``Dannet: A one-stage domain adaptation network for unsupervised nighttime semantic segmentation,'' in \emph{Proceedings of the IEEE/CVF Conference on Computer Vision and Pattern Recognition}, 2021, pp. 15\,769--15\,778.

\bibitem{chen2018domain}
Y.~Chen, W.~Li, C.~Sakaridis, D.~Dai, and L.~Van~Gool, ``Domain adaptive faster r-cnn for object detection in the wild,'' in \emph{Proceedings of the IEEE conference on computer vision and pattern recognition}, 2018, pp. 3339--3348.

\bibitem{ciampi2021domain}
L.~Ciampi, C.~Santiago, J.~P. Costeira, C.~Gennaro, G.~Amato \emph{et~al.}, ``Domain adaptation for traffic density estimation.'' in \emph{VISIGRAPP (5: VISAPP)}, 2021, pp. 185--195.

\bibitem{5288526}
S.~J. Pan and Q.~Yang, ``A survey on transfer learning,'' \emph{IEEE Transactions on Knowledge and Data Engineering}, vol.~22, no.~10, pp. 1345--1359, 2010.

\bibitem{hong2024deep}
P.~P. Hong, H.~H. Khanh, N.~N. Vinh, N.~N. Trung, A.~N. Quoc, and H.~T. Ngoc, ``Deep learning-based lane-keeping assist system for self-driving cars using transfer learning and fine tuning,'' \emph{Journal of Advances in Information Technology}, vol.~15, no.~3, pp. 322--329, 2024.

\bibitem{rothmeier2024time}
T.~Rothmeier, W.~Huber, and A.~C. Knoll, ``Time to shine: Fine-tuning object detection models with synthetic adverse weather images,'' in \emph{Proceedings of the IEEE/CVF Winter Conference on Applications of Computer Vision}, 2024, pp. 4447--4456.

\bibitem{deng2021adapt}
N.~Deng, Z.~Cao, W.~Zhou, K.~Jiang, and D.~Yang, ``Adapt the driving policy to local traffic before entering the new area,'' in \emph{2021 IEEE International Intelligent Transportation Systems Conference (ITSC)}.\hskip 1em plus 0.5em minus 0.4em\relax IEEE, 2021, pp. 798--803.

\bibitem{kim2017interpretable}
J.~Kim and J.~Canny, ``Interpretable learning for self-driving cars by visualizing causal attention,'' in \emph{Proceedings of the IEEE international conference on computer vision}, 2017, pp. 2942--2950.

\bibitem{zhou2016learning}
B.~Zhou, A.~Khosla, A.~Lapedriza, A.~Oliva, and A.~Torralba, ``Learning deep features for discriminative localization,'' in \emph{Proceedings of the IEEE conference on computer vision and pattern recognition}, 2016, pp. 2921--2929.

\bibitem{selvaraju2017grad}
R.~R. Selvaraju, M.~Cogswell, A.~Das, R.~Vedantam, D.~Parikh, and D.~Batra, ``Grad-cam: Visual explanations from deep networks via gradient-based localization,'' in \emph{Proceedings of the IEEE international conference on computer vision}, 2017, pp. 618--626.

\bibitem{sundararajan2017axiomatic}
M.~Sundararajan, A.~Taly, and Q.~Yan, ``Axiomatic attribution for deep networks,'' in \emph{International conference on machine learning}.\hskip 1em plus 0.5em minus 0.4em\relax PMLR, 2017, pp. 3319--3328.

\bibitem{zhao2024more}
S.~Zhao, H.~Chen, H.~Huang, P.~Xu, and G.~Ding, ``More is better: Deep domain adaptation with multiple sources,'' \emph{arXiv preprint arXiv:2405.00749}, 2024.

\bibitem{doshi2009roles}
A.~Doshi and M.~M. Trivedi, ``On the roles of eye gaze and head dynamics in predicting driver's intent to change lanes,'' \emph{IEEE Transactions on Intelligent Transportation Systems}, vol.~10, no.~3, pp. 453--462, 2009.

\bibitem{droste2020unified}
R.~Droste, J.~Jiao, and J.~A. Noble, ``Unified image and video saliency modeling,'' in \emph{Computer Vision--ECCV 2020: 16th European Conference, Glasgow, UK, August 23--28, 2020, Proceedings, Part V 16}.\hskip 1em plus 0.5em minus 0.4em\relax Springer, 2020, pp. 419--435.

\bibitem{canny1986computational}
J.~Canny, ``A computational approach to edge detection,'' \emph{IEEE Transactions on pattern analysis and machine intelligence}, no.~6, pp. 679--698, 1986.

\end{thebibliography}

\end{document}